# Interspecies Collaboration in the Design of Visual Identity: A Case Study

Bojan Jerbić[1], Marko Švaco[1], Filip Šuligoj[1], Bojan Šekoranja[1], Josip Vidaković[1], Marija Turković[1], Mihaela Lekić[1], Borjan Pavlek[2], Bruno Bolfan[2], Davor Bruketa[2], Dina Borošić[2], Barbara Bušić[2]

[1] Faculty of Mechanical Engineering and Naval Architecture, University of Zagreb
[2] Bruketa&Žinić&Grey

Correspondence: bojan.jerbic@fsb.hr
https://crta.fsb.hr/

**ABSTRACT**

Design usually relies on human ingenuity, but the past decade has seen the field's toolbox expanding to Artificial Intelligence (AI) and its adjacent methods, making room for hybrid, algorithmic creations. This article aims to substantiate the concept of interspecies collaboration – that of natural and artificial intelligence – in the active co-creation of a visual identity, describing a case study of the Regional Center of Excellence for Robotic Technology (CRTA) which opened on 750 m$^2$ in June 2021 within the University of Zagreb. The visual identity of the Center comprises three separately devised elements, each representative of the human-AI relationship and embedded in the institution's logo. Firstly, the letter "C" (from the CRTA acronym) was created using a Gaussian Mixture Model (GMM) applied to (x, y) coordinates that the neurosurgical robot RONNA, CRTA's flagship innovation, generated when hand-guided by a human operator. The second shape of the letter "C" was created by using the same (x, y) coordinates as inputs fed to a neural network whose goal was to output letters in a novel, AI-generated typography. A basic feedforward back-propagating neural network with two hidden layers was chosen for the task. The final and third design element was a trajectory the robot RONNA makes when performing a brain biopsy. As CRTA embodies a state-of-the-art venue for robotics research, the 'interspecies' approach was used to accentuate the importance of human-robot collaboration which is at the core of the newly opened Center, illustrating the potential of reciprocal and amicable relationship that humans could have with technology.

• Applied computing → Arts and humanities

*Interspecies Collaboration in the Design of Visual Identity: A Case Study*

## 1 INTRODUCTION

The main goal of the University of Zagreb's CRTA, the Regional Center of Excellence for Robotic Technology [Fig. 1.], is to establish a reference center for research, development, and educational activities in the field of robotics and artificial intelligence. Activities of the Center are focused on research and development of advanced robotic applications, especially in industry and medicine, but also in other areas of human activity where traditional automation and human labor should be replaced by autonomous and intelligent systems. An important step in such a venture is to enable all stakeholders of the robotics revolution, i.e., the general public and not exclusively scientists, to participate in the shaping of a new society that is to stand in close interaction with technology.

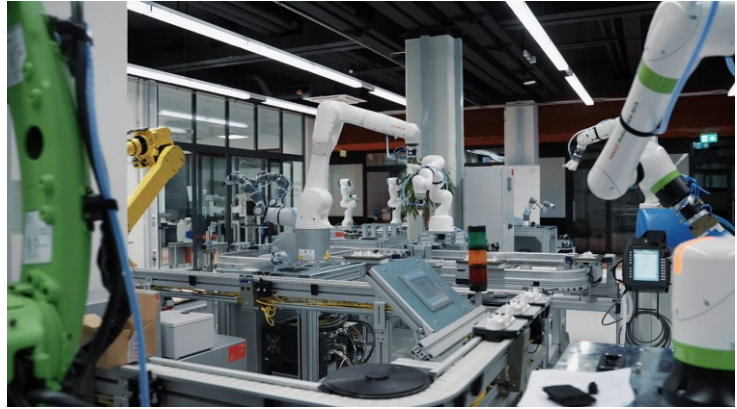

**Fig. 1. A view of collaborative robotic systems next to assembly lines in the Laboratory for Autonomous Systems, one of the three research laboratories at CRTA**

RONNA [Fig. 2.], a robot designed to perform brain surgery, was the 'co-creator' of the new visual identity for the Center. RONNA is a project successfully realized within the Faculty of Mechanical Engineering and Naval Architecture, University of Zagreb, which became a part of the Faculty and University identity. The clinical application of RONNA is intended for minimally invasive stereotactic procedures such as biopsies, stereoelectroencephalography, epilepsy surgeries, deep brain stimulation, and tumor resections [1, 2]. RONNA has performed numerous medical procedures since its first application in March 2016 in University Hospital Dubrava, one of Croatia's largest hospitals, where it has been undergoing clinical trials ever since. Although proven in neurosurgery, RONNA had to 'learn' how to write and interpret letters, a building block of the Center's visual identity.

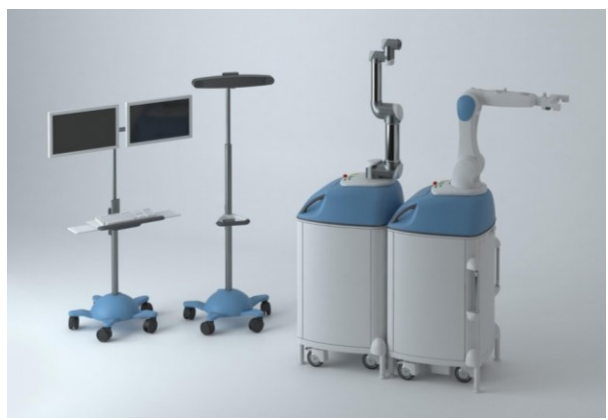

**Fig. 2. The 3rd generation of the robotic system RONNA with an assistant (left) and master robot (right)**

## 2   RELATED WORK

Since the 2010s, hundreds of art collectives had sprung out of the idea of AI-generated art, a notable commercial example being the Obvious collective from Paris [3]. Obvious relied on Robbie Barrat's code for Generative Adversarial Networks (GAN), a type of neural network architecture, to 'paint' portraits of the fictional Belamy family after being trained on more than 15,000 18th-century-style portraits and later on to sell one at Christie's, an NYC auction house, for $432,500. Questions of who gets credit for AI-based art have already been raised [4, 5] – man or machine, the engineer or the artist – hinting at the substantial human effort needed to structure the underlying databases and instructions for such seemingly autonomous systems to work, and to the interdependence of that AI-art communities have with the pioneering work of computer scientists. Though GANs are the current 'go-to' choice for most visual artists, some initiatives explored the idea of physical, robotic agents participating in the creative process. Roomba Paintings by Bobby Zokaites (2006) [6] showed how household robots could be transformed into painters by 'replacing their dust collectors with paint reservoirs'.

Apart from art, the design and communication realms have experienced a steady flow of AI innovations as well. In 2017, seven million jars of Nutella were designed by an algorithm which pulled from a database of dozens of patterns and colors to create unique packaging designs. Nutella's manufacturer Ferrero titled the project Nutella Unica [7] and worked on it with the Ogilvy & Mather Italia advertising agency. The limited edition of Nutella's label came in thousands of pattern and color combinations, each stamped with its unique code so it could be authenticated by collectors. Airbnb, the home-sharing company, developed a machine learning system [8] that can classify hand-drawn design components and turn sketches into prototypes with engineering specifications – almost instantaneously. In 2018, Lexus, the luxury division of Toyota, released an advert that had been scripted entirely by an artificial intelligence algorithm and shot by an Oscar-winning director Kevin Macdonald [9]. The algorithm was trained on 15 years of Cannes Lions International award-winning automotive ads, coupled with other data points on consumer behavior. A minute-long film is telling a story about a 'Takumi', an honorary title for a Lexus master craftsman, who completes the new Lexus ES car, which, as soon as it is released into the world, is facing an imminent crash. At a deciding moment, the car's automatic emergency braking system is activated, demonstrating the utility of intelligent features built into the car and leading the inhouse PR experts to conclude, *'This world-first collaboration between AI and a renowned craftsman (is) aimed to test the boundaries of how humans and machines can work together in perfect harmony, exploring the importance of intuition in the relationship between the two.'* The logo design process has also been impacted, though in limited capacity, because of the particular, aggravating properties of logos themselves. Unlike images, logos are hard to label since they contain a very low number of properties that are at the same time rich in meaning and suggestion [10], as emphasized by Sage et al. who built a logo synthesis GAN algorithm trained on more than 600,000 web-scraped logos.

Most of the aforementioned initiatives were carried out *in silico*, or in the virtual space, without interaction with elements in the natural environment such as humans. These projects framed the importance of AI-based algorithms in art and design as (quasi)independent and isolated creators, though some of them pointed to the symbiotic potential that AI could realize only when actively collaborating with humans – in 'interspecies' collaboration – 'on equal terms', as working colleagues taking into consideration the other side's feedback. The term 'interspecies collaboration' was previously used almost entirely in the context of collaboration of humans and animals or plants [11, 12] while this article aims to stretch that definition to cover AI-based algorithms which are proving to have ever-increasing agency, forming a sort of 'species' with an established relationship to their evolutionary predecessors, i.e., humans.

## 3 METHOD

The approach to devising a visual identity for CRTA was to form a graphic solution consisting of three elements – each of which, in its own way, connects man and artificial intelligence. The components were chosen in collaboration with the Bruketa&Žinić&Grey agency and were as follows: 1. a hand-guided letter "C" (the first letter of the CRTA acronym) realized through the movement of RONNA's robotic arm; 2. an artificially generated letter "C" realized as a neural network output; 3. a brain surgery trajectory of the robot RONNA [Fig. 3.].

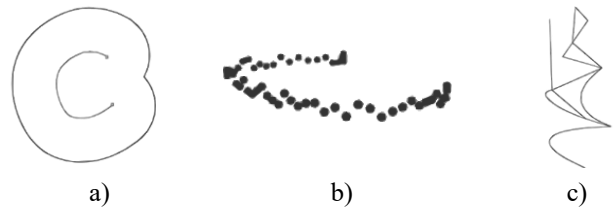

a)     b)     c)

**Fig. 3. A tripartite strategy for the visual identity of CRTA**
**a) A hand-guided letter "C", b) An artificially-generated letter "C",**
**c) A brain surgery trajectory by RONNA**

### 3.1  A hand-guided letter "C"

The staff at CRTA, professors and students, each hand-guided the RONNA assistant robot to write the letter "C" in their own gestures and on a whiteboard [Fig. 4.], resulting in nine depictions of the letter, each unique to the person hand-guiding the robot and thus making the CRTA staff a part of the visual identity of the institution they would soon be actively involved with. The robot recorded (x, y) coordinates generated during its movement through space [Table 1.] with a fixed step of 10 Hz.

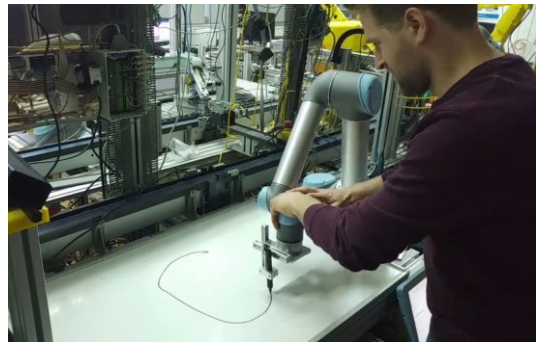

**Fig. 4. A member of CRTA staff hand-guiding the robot**

```
-0.253, 0.341     -0.255, 0.345     -0.307, 0.424     -0.515, 0.282
-0.255, 0.345     -0.255, 0.346     -0.326, 0.426     -0.527, 0.256
-0.255, 0.345     -0.257, 0.351     -0.348, 0.426     -0.537, 0.231
-0.255, 0.345     -0.260, 0.357     -0.372, 0.422     -0.545, 0.205
-0.255, 0.345     -0.263, 0.367     -0.394, 0.414     -0.550, 0.183
-0.255, 0.345     -0.268, 0.378     -0.419, 0.403     -0.548, 0.161
-0.255, 0.345     -0.271, 0.387     -0.443, 0.385     -0.540, 0.143
-0.255, 0.345     -0.276, 0.398     -0.462, 0.366     -0.527, 0.130
-0.255, 0.345     -0.281, 0.409     -0.481, 0.340     -0.515, 0.122
-0.255, 0.345     -0.291, 0.418     -0.498, 0.312     -0.505, 0.119
       …                 …                 …
```

**Table 1.  Example of robot coordinates for a single letter "C"**

A Gaussian Mixture Model (GMM) was applied to the hand-guided (x, y) coordinates of the robot trajectory, as described and implemented by Vidaković et al. [13]. The resulting effect was a smooth and vectorized curve of the letters written through human-robot collaboration. The generalized letter "C" was used as a baseline for further graphic interventions by the human designer and ultimately formed the dominant part of the visual identity of the whole Center. The figure below [Fig. 5.] shows a GMM used for describing two-dimensional spatial curves with 10 normal distributions. Each normal distribution represents a field that 'attracts' the trajectories. A generalized curve was therefore given by approximation, using the coordinates of centers of normal distributions, and finally vectorized [Fig. 6.].

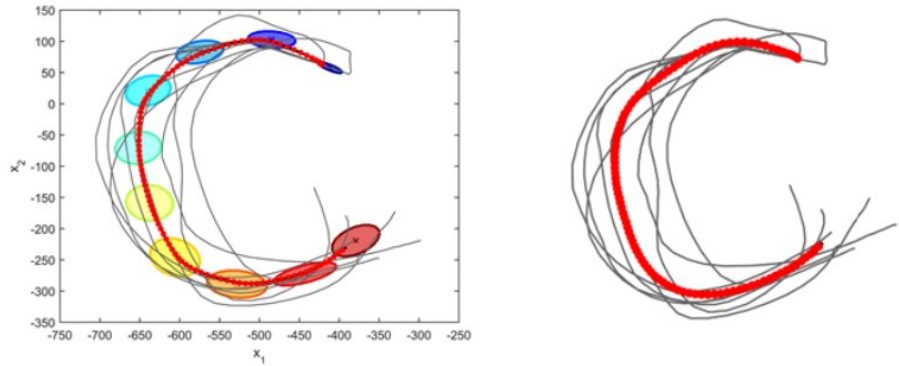

Fig. 5. Curve generalization using GMM

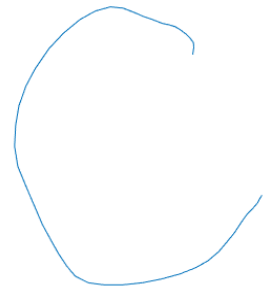

Fig. 6. Example of a vectorized curve of a single hand-guided letter
derived from nine other letters

### 3.2 An artificially-generated letter "C"

All hand-guided motifs were analyzed through an artificial neural network [14, 15] that gave its interpretation of the motifs by creating new representations of the input letters. By overlapping the human and artificially created motifs, a unique graphic art was generated – specific to the time, space, humans, and robots that participated in the creative process. The goal of the neural network was to learn to create different kinds of letters. However, since there were only a few input values (the dataset consisted of only nine letters generated during nine human-robot collaboration instances), the algorithm was constructed in a way that makes combinations of input and output pairs and in that way it was artificially enlarged to 56 input-output pairs in total [Table 2.]. The input-output pairs of images with the same indices (n-n) were excluded from the dataset as the network wouldn't be able to learn from these.

| (Input Image – Output Image) pairs ||||||||
|---|---|---|---|---|---|---|---|
| 1 - 2 | 2 - 1 | 3 - 1 | 4 - 1 | 5 - 1 | 6 - 1 | 7 - 1 | 8 - 1 |
| 1 - 3 | 2 - 3 | 3 - 2 | 4 - 2 | 5 - 2 | 6 - 2 | 7 - 2 | 8 - 2 |
| 1 - 4 | 2 - 4 | 3 - 4 | 4 - 3 | 5 - 3 | 6 - 3 | 7 - 3 | 8 - 3 |
| 1 - 5 | 2 - 5 | 3 - 5 | 4 - 5 | 5 - 4 | 6 - 4 | 7 - 4 | 8 - 4 |
| 1 - 6 | 2 - 6 | 3 - 6 | 4 - 6 | 5 - 6 | 6 - 5 | 7 - 5 | 8 - 5 |
| 1 - 7 | 2 - 7 | 3 - 7 | 4 - 7 | 5 - 7 | 6 - 7 | 7 - 6 | 8 - 6 |
| 1 - 8 | 2 - 8 | 3 - 8 | 4 - 8 | 5 - 8 | 6 - 8 | 7 - 8 | 8 - 7 |

**Table 2. Artificial enlargement of the input dataset using nine letters to form 56 input-output pairs**

The dataset was divided into two groups – the training set which had different combinations of input and output pairs, and the test set. The input of a letter "C" written in one way would result in a different font of the same letter. The test set consisted of only one image, because the training set needed to be as large as possible and, in theory, only one output image needed to be forwarded to the graphic designer for final adjustments. The neural network was a feedforward back-propagating network with two hidden layers. The number of neurons in each layer was 10. The network had a learning rate of 0.0001 and the number of iterations was 10,000. The hyperbolic tangent activation function was used, and the optimization function was stochastic gradient descent. The cost function used for training was the following:

$$\text{Train Cost} = \frac{\sum (y_{predicted} - y_{actual})^2}{2}$$

Weights (w) and biases (b) were initialized as random values from a uniform distribution in the range of [0, 1] for weights and [0, 0.1] for biases. The formulas for each layer were as follows:

1. `x == input`                                                [first layer]
2. `a1 = tanh (input x w1 + b1)`     [1st hidden layer]
3. `a2 = tanh (a1 x w2 + b2)`        [2nd hidden layer]
4. `output == y = a2 x w3 + b3`      [last layer]

The input and output values were the robot coordinates in 2D space (x, y), i.e., the horizontal whiteboard. Since all the inputs and outputs (x, y) were in the range of [-1, 1], there was no need to normalize them. The artificially generated letter "C" representing a dotted trajectory [Fig. 7.] was then handed to the designers at the Bruketa&Žinić&Grey agency for final adjustments. The "C" was extruded randomly to form the z-axis and eventually turned into a new 3D object which could be viewed in 360°. The neural network output was then rearranged on the canvas of the logos in various configurations, therefore embedding the AI-generated typography into the core design of the Center.

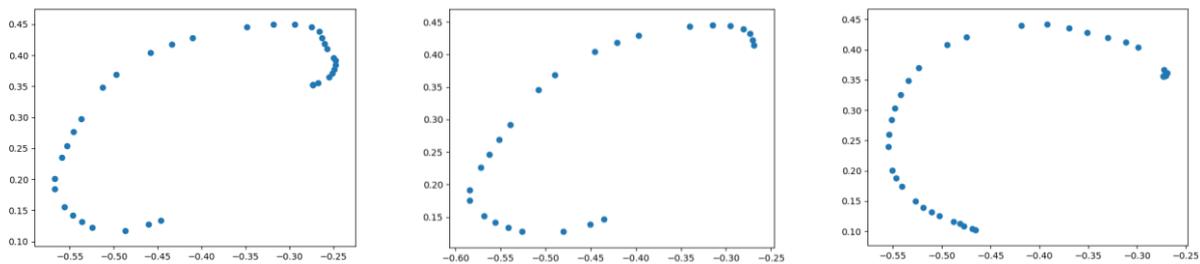

**Fig. 7. Examples of neural network outputs of the letter "C" in a novel, AI typography**

### 3.3 A brain surgery trajectory by RONNA, CRTA's flagship robotic innovation

A neurosurgery trajectory [Fig. 8.] was generated by recording points of RONNA performing a patient localization and brain biopsy procedure with a fixed time step of 10 Hz. The points were interpolated and finally formed a continuous trajectory.

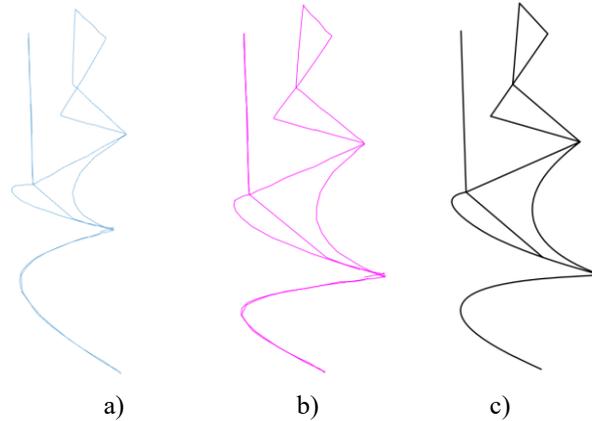

a)      b)      c)

**Fig. 8. Neurosurgery trajectory of the robot RONNA**
**a) 3D trajectory of brain surgery by RONNA, b) 2D vector representation**
**of the trajectory, c) Simplified vector representation of the trajectory**

The final design elements [Fig. 6., Fig. 7, Fig. 8.] were then assembled to form the definite visual identity of CRTA [Fig. 10., Fig. 11.], an endeavor which required various post-processing steps executed in parallel [Fig. 9.].

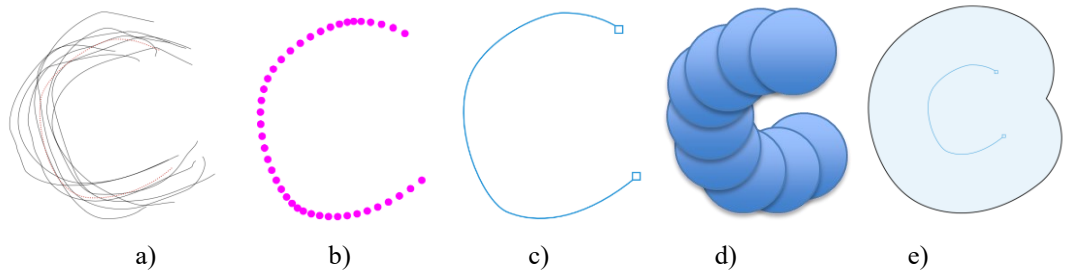

a)      b)      c)      d)      e)

**Fig. 9. Different shapes of letters specific to the way they were created**
**a) Hand-guided letters, b) Neural network input, c) Vectorized generalization of hand-guided robot trajectories, d) and e) Interventions by a human designer**

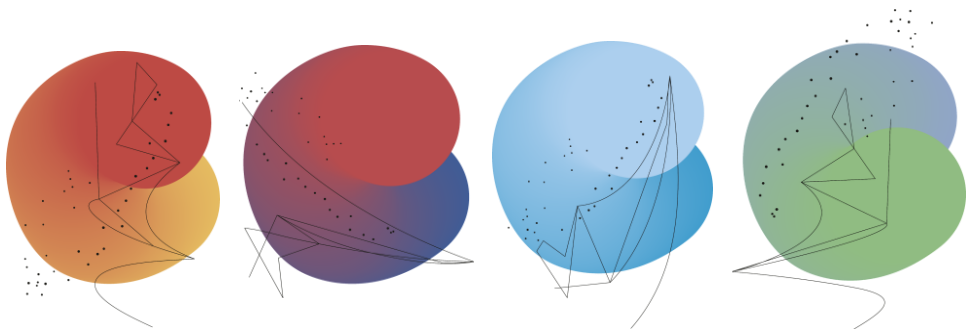

**Fig. 10. Variations of the final visual identity of CRTA**

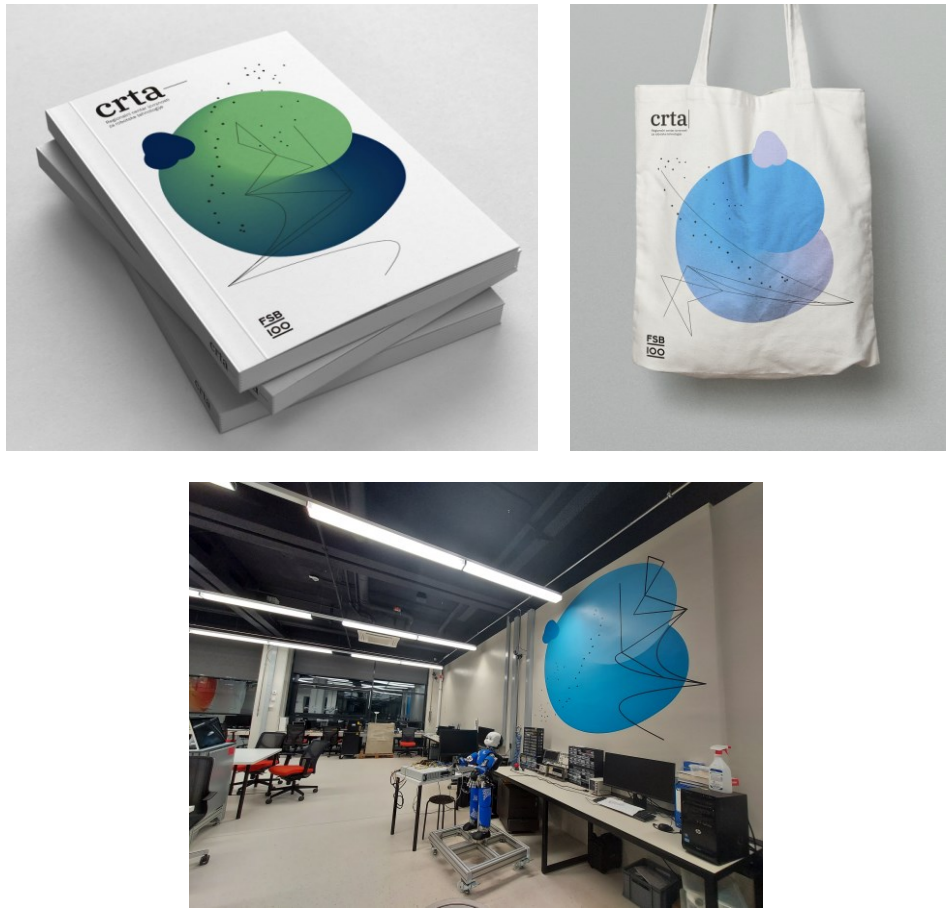

**Fig. 11. Applications of CRTA's symbolic logo in office stationery
and the physical space of the Center**

## 4   CONCLUSION

We demonstrated how neural networks and statistical methods could be applied to produce a unique, authentic and highly representative visual identity of an institution in close, 'interspecies' collaboration with humans. Three design elements were created: a vectorized curve based on hand-guided letters "C", a reinterpretation of the letter "C" in a novel, AI typography, and a neurosurgery trajectory of the robot RONNA. The incorporation of various artificially generated design components into the final design was crucial in communicating all important aspects of the newly opened Regional Center of Excellence for Robotic Technology (CRTA). We believe this case study could serve as a valuable precedent for branding initiatives aimed at leveraging the immense expressive reserves of statistical and AI tools for designers.

## 5   ACKNOWLEDGEMENTS

The authors would like to acknowledge the support of the European Union's European Regional Development Fund which financed the project CRTA.

*Interspecies Collaboration in the Design of Visual Identity: A Case Study*# REFERENCES

[1] Jerbić, B., Švaco, M., Chudy, D., Šekoranja, B., Šuligoj, F., Vidaković, J., Dlaka, D., Vitez, N., Župančić, I., Drobilo, L., Turković, M., Žgaljić, A., Kajtazi, M., & Stiperski, I. (2020). RONNA G4—Robotic Neuronavigation: A Novel Robotic Navigation Device for Stereotactic Neurosurgery. Handbook of Robotic and Image-Guided Surgery, 599–625. https://doi.org/10.1016/B978-0-12-814245-5.00035-9

[2] Dlaka, D., Švaco, M., Chudy, D., Jerbić, B., Šekoranja, B., Šuligoj, F., Vidaković, J., Romić, D., & Raguž, M. (2021). Frameless stereotactic brain biopsy: A prospective study on robot-assisted brain biopsies performed on 32 patients by using the RONNA G4 system. The International Journal of Medical Robotics and Computer Assisted Surgery, 17(3), e2245. https://doi.org/10.1002/RCS.2245

[3] Christie's (2018, accessed 20 January 2022). Is artificial intelligence set to become art's next medium? https://www.christies.com/features/a-collaboration-between-two-artists-one-human-one-a-machine-9332-1.aspx

[4] Epstein, Z., Levine, S., Rand, D. G., & Rahwan, I. (2020). Who Gets Credit for AI-Generated Art? IScience, 23(9), 101515. https://doi.org/10.1016/J.ISCI.2020.101515

[5] McCormack, J., Gifford, T., & Hutchings, P. (2019). Autonomy, Authenticity, Authorship and Intention in computer generated art. Lecture Notes in Computer Science (Including Subseries Lecture Notes in Artificial Intelligence and Lecture Notes in Bioinformatics), 11453 LNCS, 35–50. https://doi.org/10.1007/978-3-030-16667-0_3

[6] Zokaites, B. (2006, accessed 20 January 2022). Roomba Paintings: 2006. http://www.bobbyzokaites.com/roomba-paintings

[7] Aouf, R. S. for Dezeen (2017, accessed 20 January 2022. Algorithm designs seven million different jars of Nutella. https://www.dezeen.com/2017/06/01/algorithm-seven-million-different-jars-nutella-packaging-design/

[8] Benjamin Wilkins for Airbnb (2017, accessed 20 January 2022). Sketching Interfaces – Generating code from low fidelity wireframes. https://airbnb.design/sketching-interfaces/

[9] Lexus Europe Newsroom (2018, accessed 20 January 2022). https://newsroom.lexus.eu/driven-by-intuition-car-by-lexus-story-by-artificial-intelligence-camera-by-oscar-winning-director/

[10] Sage, A., Agustsson, E., Timofte, R., & van Gool, L. (2017). Logo Synthesis and Manipulation with Clustered Generative Adversarial Networks. Proceedings of the IEEE Computer Society Conference on Computer Vision and Pattern Recognition, 5879–5888. https://doi.org/10.1109/CVPR.2018.00616

[11] Jevbratt, L. (2009). Interspecies Collaboration - Making Art Together with Nonhuman Animals. Presented at the Minding Animals conference in Newcastle, Australia in July 2009. http://jevbratt.com/home_texts.html

[12] Fischer, D. (2020). Art Between Species: Two Case Studies of Animal's Agency in Interspecies Art (pp. 67–92). Leiden University Centre for the Arts in Society. https://hdl.handle.net/1887/123086

[13] Vidaković, J., Jerbić, B., Šekoranja, B., Švaco, M., & Šuligoj, F. (2019). Learning from Demonstration Based on a Classification of Task Parameters and Trajectory Optimization. Journal of Intelligent & Robotic Systems 2019 99:2, 99(2), 261–275. https://doi.org/10.1007/S10846-019-01101-2

[14] Svozil, D., Kvasnička, V., & Pospíchal, J. (1997). Introduction to multi-layer feed-forward neural networks. Chemometrics and Intelligent Laboratory Systems, 39(1), 43–62. https://doi.org/10.1016/S0169-7439(97)00061-0

[15] Hayou, S., Doucet, A., & Rousseau, J. (2019). On the Impact of the Activation function on Deep Neural Networks Training (pp. 2672–2680). PMLR. https://proceedings.mlr.press/v97/hayou19a.html
9